%% file: CRAB.tex
\documentclass[acmtog]{acmart}

\input{src/macros}

\copyrightyear{2026}
\acmYear{2026}
\setcopyright{cc}
\setcctype{by}
\acmConference[SA Conference Papers '26]{SIGGRAPH Asia 2026 Conference Papers}{December 01--04, 2026}{Kuala Lumpur, Malaysia}
\acmBooktitle{SIGGRAPH Asia 2026 Conference Papers (SA Conference Papers '26), December 01--04, 2026, Kuala Lumpur, Malaysia}
\acmDOI{10.1145/3829340.3842245}
\acmISBN{979-8-4007-2842-6/2026/12}

\begin{document}
\title{CRAB: Cuboid-based Representation for As-built BIM from Point Clouds}


\author{Haoran Xiong}
\orcid{0009-0005-7961-7648}
\affiliation{%
     \department{Guangdong Provincial Key Laboratory of Visual Media and Multidimensional Intelligence, College of Computer Science and Software Engineering}
	\institution{Shenzhen University}
	\country{China}	
    \city{Shenzhen}
}
\email{xhr2002me@gamil.com}

\author{Jing Wang}
\orcid{0000-0002-9792-7165}
\affiliation{%
	\institution{Shenzhen University}
	\country{China}	
    \city{Shenzhen}
}
\email{wangjinginszu@gmail.com}

\author{Shanshan Pan}
\orcid{0009-0008-8259-5234}
\affiliation{%
	\institution{Shenzhen University}
	\country{China}	
    \city{Shenzhen}
}
\email{psshappystar@gmail.com}

\author{Zihao Yan}
\orcid{0000-0003-2160-9109}
\affiliation{%
	\institution{Shenzhen University}
	\country{China}	
    \city{Shenzhen}
}
\email{mr.salingo@gmail.com}

\author{Minglun Gong}
\orcid{0000-0001-5820-5381}
\affiliation{%
	\institution{University of Guelph}
	\country{Canada}	
    \city{Guelph}
}
\email{gongml@gmail.com}

\author{Hui Huang}
\orcid{0000-0003-3212-0544}
\authornote{Corresponding author: Hui Huang (hhzhiyan@gmail.com)}
\affiliation{%
     \department{Guangdong Provincial Key Laboratory of Visual Media and Multidimensional Intelligence, College of Computer Science and Software Engineering}
	\institution{Shenzhen University}
	\country{China}	
    \city{Shenzhen}
}
\email{hhzhiyan@gmail.com}

\begin{teaserfigure}
  \centering
  \includegraphics[width=0.97\linewidth]{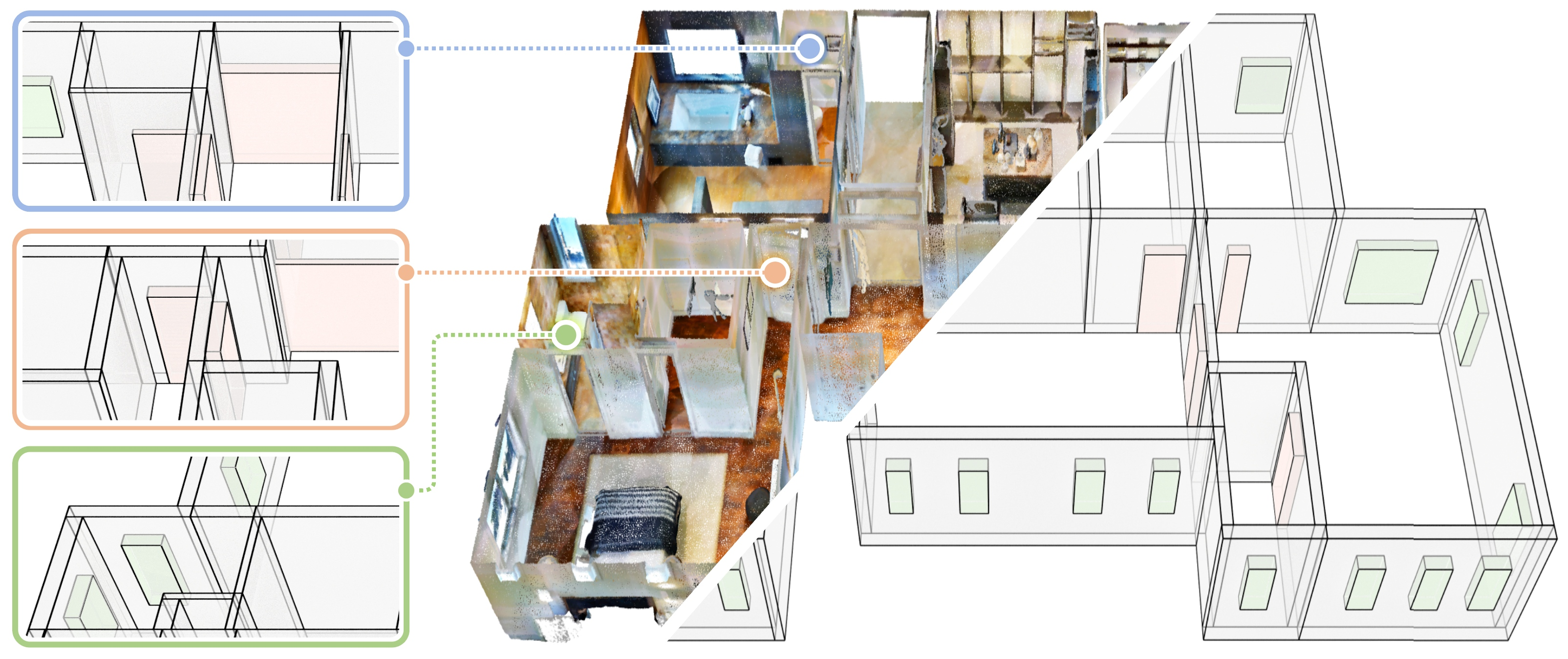}
\caption{
\emph{CRAB} converts cluttered and incomplete indoor scans into compact, BIM-ready cuboid assemblies.
By representing \hr{walls, doors, windows, beams, and columns} as semantic cuboids with explicit geometric parameters and structural relations, \emph{CRAB} supports accurate reconstruction, local editability, and downstream BIM conversion.
The zoomed regions highlight challenging structures, including dense room partitions, non-uniform wall thicknesses, corner openings, and complex interior cores, while demonstrating precise wall-thickness recovery, consistent door-wall junctions, and robust reconstruction.
}
  \Description{}
  \label{fig:teaser}
\end{teaserfigure}


\input{src/00-abstract}


%
%


%
%


\maketitle

\input{src/01-intro}
\input{src/02-rw}
\input{src/03-dataset}

\input{src/04-method}

\input{src/05-results}
\input{src/06-future}

\begin{acks}
This work was supported in part by Guangdong S\&T Program (2024B0101050004), ICFCRT (W2441020), NSFC(62402323), Shenzhen Science and Technology Program (KQTD20210811090044003, KJZD20240903100022028), and Scientific Development Fund from Guangdong Provincial Key Laboratory of Visual Media and Multidimensional Intelligence.
\end{acks}

\bibliographystyle{ACM-Reference-Format}
\bibliography{CRAB}

\input{figs/fig_compare_sota}
\input{figs/fig_compare_real}

\end{document}

%% file: src/macros.tex
\usepackage{color}
\usepackage{xcolor}
\usepackage{graphicx}
\usepackage{amsmath, amsthm, amsfonts, amssymb}
\usepackage{textcomp}
\usepackage{multirow}
\usepackage{subfig}
\usepackage{booktabs} 
\usepackage[ruled]{algorithm2e} 
\usepackage{cleveref}
\usepackage{bm}
\usepackage{enumitem}

\SetAlFnt{\small}
\SetAlCapFnt{\small}
\SetAlCapNameFnt{\small}
\SetAlCapHSkip{0pt}

\newcommand{\eg}{e.g.}

\definecolor{green}{rgb}{0, 0.5, 0}
\definecolor{orange}{rgb}{0.6, 0.3, 0.1}
\definecolor{red}{rgb}{1.0, 0.0, 0.0}
\definecolor{teal}{rgb}{0.0, 0.4, 0.4}
\definecolor{purple}{rgb}{0.65,0,0.65}
\definecolor{saffron}{rgb}{0.95,0.75,0.2}
\definecolor{turquoise}{rgb}{0.0,0.5,0.5}
\definecolor{brown}{rgb}{0.5, 0.16, 0.16}
\definecolor{brickred}{rgb}{.6, .2 .1}
\definecolor{coral}{rgb}{1,0.45,0.33}
\definecolor{newcolor}{rgb}{.8,.349,.1}

\definecolor{hh}{RGB}{255,0,0}

\definecolor{ml}{RGB}{165, 42, 42}

\definecolor{wj}{RGB}{0,165,255}
\definecolor{wj2}{RGB}{0,120,200}

\newcommand{\hr}[1]{\color{black}{#1}}

\definecolor{mygray}{gray}{.9}
\definecolor{zh}{RGB}{0,125,125}

\usepackage{colortbl}

\usepackage{pifont}
\usepackage{makecell}
\usepackage{array}

\definecolor{Best}{HTML}{D6F5D6}   %
\definecolor{Second}{HTML}{E8F0FE} %

\newcommand{\bestdown}[1]{\cellcolor{Best}\textbf{#1}}
\newcommand{\seconddown}[1]{\cellcolor{Second}#1}
\newcommand{\bestup}[1]{\cellcolor{Best}\textbf{#1}}
\newcommand{\secondup}[1]{\cellcolor{Second}#1}

%% file: src/00-abstract.tex
\begin{abstract}
As-built BIM reconstruction from point clouds demands semantic, volumetric, parametric, and editable building components, beyond 2D layouts or planar abstractions. 
Yet existing scan-to-BIM pipelines often rely on hand-crafted rules, heuristic optimization, or manual correction, while learning-based methods fall short of recovering wall thickness, opening geometry, and component-level BIM structure. 
We introduce \textbf{CRAB}, a BIM-ready cuboid reconstruction framework that encodes \hr{walls, doors, windows, beams, and columns} as compact architectural cuboids with IFC-compatible semantics and parameters. 
\emph{CRAB} bridges point clouds and editable BIM by combining \textit{ProposalNet}, which predicts coarse component cuboids, with \textit{AssemblyNet}, a relation-aware Flow Matching model that refines geometry, predicts structural relationships, and enables automatic structural correction and human-guided editing.
To address limited cuboid-level supervision, we extend Structured3D with component-wise cuboid annotations and construct cuboid-based procedural scenes.
Experiments show \emph{CRAB} surpasses floor plan and plane-based alternatives in component completeness, geometric accuracy, and generalization. 
For interior wall thickness estimation, \emph{CRAB} achieves a mean error of 2.80 cm, with 79.50\% of matched walls within 3 cm.
The project page is available at \url{https://vcc.tech/research/2026/CRAB}.
\end{abstract}

%% file: src/01-intro.tex
\section{Introduction}\label{sec:intro}

3D scanning is widely used on construction sites to capture the geometric structure of real-world buildings and environments. The resulting point clouds often provide noisy, incomplete, and purely geometric surface measurements. In contrast, Building Information Modeling (BIM) represents buildings as structured collections of semantic component instances with volumetric geometry, parametric attributes, and explicit relationships\cite{zbirovsky2025cloud2bim,liu2025bimnet}. Consequently, BIM aims to reconstruct editable as-built architectural components rather than merely producing a geometric reconstruction of the scanned scene. For architectural components, this requires inferring wall thickness, wall-opening relationships, and editable instance identities that are not directly observable from raw scans.

Current methods rarely recover such BIM components directly from point clouds. 
Manual modeling remains labor-intensive, while classical Scan-to-BIM pipelines rely on multi-stage processing, including segmentation, primitive fitting, plane extraction, and topology optimization, often requiring scene-specific assumptions and carefully tuned parameters~\cite{zbirovsky2025cloud2bim,liu2025bimnet}. 
Learning-based approaches improve automation, but most focus on intermediate geometric abstractions such as floor plans, room polygons~\cite{yue2023roomformer}, or 3D planes~\cite{mao2025spatialLM}. 
\hr{Although these representations support spatial understanding, they describe
layouts or surfaces rather than explicit volumetric component instances.
They do not directly couple component geometry, limiting their
extensibility to other BIM attributes such as load-bearing status and
materials.} 
A reconstruction representation should therefore expose these attributes as primary variables, rather than deriving them from layouts or planes after prediction.

This mismatch calls for a volumetric, component-aware representation. 
Cuboid and primitive assemblies are a natural candidate because they encode compact solids with interpretable parameters and support editable 3D modeling. 
However, existing cuboid-based methods primarily target object decomposition\cite{ye2025primitiveanything}, shape generation~\cite{jones2020shapeassembly}, or furniture arrangement~\cite{feng2025casagpt}, rather than architectural reconstruction from incomplete scene-level point clouds with ambiguous component boundaries.
\hr{Moreover, their generic cuboid parameterizations are not tailored to highly
elongated architectural components. This motivates a BIM-oriented volumetric
representation that explicitly models parametric component geometry and
architectural relations.}

To this end, we introduce \textbf{CRAB} (\textbf{C}uboid-based \textbf{R}epresentation for \textbf{A}s-built \textbf{B}IM), a BIM-ready proposal-and-assembly framework for indoor point clouds\hr{. CRAB is designed precisely to move
beyond layout and plane abstractions toward volumetric BIM components,
thereby demonstrating the value of a BIM-oriented cuboid representation for
AI-based Scan-to-BIM. It reconstructs walls, doors, windows, beams, and
columns in single-floor scenes and represents each scene as centerline-based
semantic cuboids and explicit topological relationships, jointly encoding
parametric component geometry, semantics, wall adjacency, and opening-wall
affiliation.}

\hr{To recover these cuboids and relationships from noisy and incomplete point
clouds, CRAB adopts a two-stage design.} \textit{ProposalNet} first generates coarse semantic cuboids from point cloud features, while \textit{AssemblyNet} further refines their geometry and predicts inter-component relationships through relation-aware Flow Matching. This refinement improves structural consistency while preserving editable cuboid parameters for user-guided modification. The resulting components can be directly instantiated as Industry Foundation Classes (IFC) elements, enabling compatibility with the open BIM data-exchange standard.

A further challenge is the lack of cuboid-level supervision for scan-to-BIM learning. 
Existing indoor datasets typically provide floor plans, meshes, semantic labels, or layout annotations, but lack component-wise volumetric cuboids aligned with BIM elements. 
To support training and evaluation, we extend the public Structured3D~\cite{jia2020structured3d} dataset with cuboid annotations for architectural components, and further construct a cuboid-based synthetic dataset with procedurally generated scenes and simulated point cloud acquisition.

In summary, our contributions are threefold:

\begin{itemize}[noitemsep,topsep=0pt,leftmargin=0em,itemindent=2em]
    \item We recast scan-based indoor BIM reconstruction as semantic cuboid assembly, using centerline-based cuboid parameters to explicitly model component semantics, wall thickness, opening height, and editable volumetric geometry.
    
    \item We construct \emph{CrabSet}, a scalable cuboid-level supervision dataset for Scan-to-BIM, by enriching Structured3D with component-wise cuboid annotations and procedurally synthesizing diverse architectural scenes with simulated point-cloud acquisition.
    
    \item We introduce \emph{CRAB}, a proposal-and-assembly framework that couples \textit{ProposalNet} for semantic cuboid generation with \textit{AssemblyNet} for relation-aware flow-matching refinement, producing accurate, structurally coherent, and editable BIM-ready models.

\end{itemize}

%% file: src/02-rw.tex
\section{Related Work}
\label{sec:rw}

\subsection{Structured Indoor Modeling}
{Indoor structural modeling is often cast as floor plan or layout estimation. 
Existing methods reconstruct 2D graphs, room polygons, or polygonal loops from RGB-D scans and point-density maps~\cite{chen2019floorsp,chen2022heat,stekovic2021montefloor,yue2023roomformer}, with recent variants improving polygon generation, room representation, graph reasoning, segmentation, and edge continuity~\cite{chen2023polydiffuse,xu2024frinet,liu2024polyroom,sun2025polygraph,ye2025floorsam,liu2025cage}. 
While compact, these floorplan-oriented outputs do not directly specify wall thickness, vertical extent, opening height, or component volume, which are required by CAD/BIM workflows.}

Recent 3D structured modeling methods move beyond 2D layouts by predicting walls, openings, oriented boxes, or parametric layout quads from images or point clouds~\cite{avetisyan2024scenescript,mao2025spatialLM,chen2022pqtransformer,bieri2025houselayout3d}.
\hr{For planar representations, recovering volumetric semantic component
instances from predicted paired parallel planes requires additional plane
pairing and instance association to infer wall thickness, opening affiliation,
and component identity. Such associations can be ambiguous when a single
plane spans multiple walls, limiting extensions to instance-specific BIM
attributes such as materials and load-bearing status.}
We instead reconstruct cuboid assemblies for walls, doors, windows, beams, and columns, with explicit centerline, thickness, height.

\subsection{BIM-Ready Cuboid Reconstruction}
{Scan-to-BIM converts sensed geometry into editable building models. 
Existing systems recover walls, slabs, openings, rooms, and IFC-compatible outputs from point clouds~\cite{murali2017indoor,ochmann2019volumetric,zbirovsky2025cloud2bim}, and BIMNet highlights the need for domain-specific data and BIM-aware geometric and topological evaluation~\cite{liu2025bimnet}. 
Yet many pipelines still rely on geometric rules, segmentation, optimization, or software-specific workflows, making them sensitive to clutter, occlusion, and incomplete scans~\cite{wu2023flkpp,han2023floorusg,wong2025imagebased}.}
\hr{Moreover, they primarily recover final component geometry without explicitly modeling relations, limiting coordinated refinement.}

Cuboid and primitive assemblies offer compact, editable structures. 
ShapeAssembly represents shapes as programs over cuboid parts~\cite{jones2020shapeassembly}, while recent primitive and procedural methods learn structured decompositions or editable programs from data~\cite{ye2025primitiveanything,raji2026proc3d}. 
CASAGPT uses cuboid primitives for indoor scene synthesis to reduce mesh-level collisions~\cite{feng2025casagpt}.
\hr{Existing methods mainly target object decomposition, shape generation, or
furnished scene synthesis rather than architectural reconstruction from
incomplete scene-level point clouds. They typically use generic cuboids
parameterized by center, size, and orientation, which are not tailored to
highly elongated architectural components. CRAB instead adopts
centerline-based semantic cuboids and explicit relations, supporting coordinated refinement and
editable component reconstruction.}

\subsection{Local Structural Refinement}
Recent LLM/VLM agents have moved generation and editing from one-shot prompting to closed-loop planning, tool use, evaluation, and revision. 
SceneCraft refines Blender-code scenes with rendered feedback~\cite{hu2024scenecraft}, and VIGA reconstructs editable graphics programs through a code-render-inspect loop~\cite{yin2026viga}. 
Similar feedback-driven loops appear in visual editing~\cite{yao2026photoagent}, while BIM agents invoke APIs, manipulate IFC data, or operate authoring software~\cite{du2026text2bim,nithyanantham2025mcp4ifc,deng2025bimgent}. 
Xie et al.~\cite{xie2025scenescriptrefine} further show the value of local correction for missed or misplaced 3D layout elements.
Motivated by these local correction paradigms, our cuboid representation and \emph{AssemblyNet} support user-guided refinement, enabling controllable local edits and global correction for improved structural consistency.

%% file: src/03-dataset.tex
\section{CrabSet Construction}\label{sec:dataset}

Learning BIM-ready cuboid reconstruction requires component-level volumetric supervision, including the geometry and relations of walls, doors, windows, and other elements. 
However, as summarized in Table~\ref{tab:tab_dataset_summary}, most existing indoor datasets are designed for scene understanding, embodied AI, or layout estimation, and mainly provide 2D floor plans, room layouts, semantic labels, or structured planar annotations. 
Only a few datasets contain explicit volumetric building components, and they are limited in scale or scene diversity. 

To this end, we construct \emph{CrabSet}, a cuboid-level dataset for indoor BIM reconstruction from point clouds. 
Each sample contains an input point cloud, component-wise volumetric cuboid annotations, and pairwise relations among architectural components. 
Each cuboid records its semantic category, instance identity, position, orientation, and dimensions.
As shown in Fig.~\ref{fig:dataset}, \emph{CrabSet} combines component-wise cuboid annotations derived from Structured3D with procedurally synthesized scenes, covering both residential layouts and broader architectural configurations. For synthetic scenes, point clouds are generated through virtual scanning.

\input{tab/tab_dataset_summary}
\input{figs/fig_dataset}

\paragraph{S3D-derived.}
This subset is derived from Structured3D, which provides diverse residential layouts with room partitions, openings, non-Manhattan structures, and occlusions. 
Using its planar faces, normals, semantic labels, and outer contours, we convert walls, doors, and windows into component-wise volumetric cuboids. 
This subset anchors \emph{CrabSet} to a public residential benchmark.

\paragraph{Boundary-varied.}
This subset increases local geometric diversity by varying outer contours, wall thicknesses, and non-Manhattan wall configurations. 
It encourages the model to recover accurate cuboid geometry and wall thickness.

\paragraph{Scale-varied.}
This subset expands spatial scale and component complexity through large architectural spaces with service shafts, columns, and multi-layer window configurations. 
It improves the model's ability to handle dense non-residential scenes with large variations in component size and spatial extent.

\paragraph{Corridor-centered.}
This subset focuses on circulation-driven layouts, including straight, bent, notched, H-shaped, T-shaped, and ring-like corridors. 
Rooms and openings are arranged along corridor boundaries, covering large building configurations that are rarely represented in residential datasets.


%% file: tab/tab_dataset_summary.tex
\begin{table}[!t]
\centering
\caption{
Structural supervision in indoor building datasets.
We compare whether each dataset provides 2D floor plan, structured 3D plane, and volumetric component-level annotations for BIM-oriented reconstruction.
}
\label{tab:tab_dataset_summary}
\setlength{\tabcolsep}{4.2pt}
\renewcommand{\arraystretch}{1.08}
\resizebox{\linewidth}{!}{
\begin{tabular}{lccccc}
\toprule
Dataset & Source & Scale & 2D Floor Plan & 3D Plane & Volumetric Comp. \\
\midrule
ISPRS Benchmark~\citeyearpar{khoshelham2017isprs} & Real-world & 6       & $\times$     & $\times$     & $\checkmark$ \\
Structured3D~\citeyearpar{jia2020structured3d}    & Synthetic  & 3,500   & $\checkmark$ & $\checkmark$ & $\times$ \\
SceneCAD~\citeyearpar{avetisyan2020scenecad}      & Real-world & 1,151   & $\checkmark$ & $\times$     & $\times$ \\
ProcTHOR~\citeyearpar{matt2022procthor}           & Synthetic  & 12,000  & $\checkmark$ & $\times$     & $\times$ \\
HSSD~\citeyearpar{khanna2024hssd}                 & Synthetic  & 211     & $\checkmark$ & $\times$     & $\times$ \\
ASE~\citeyearpar{avetisyan2024scenescript}        & Synthetic  & 100,000 & $\checkmark$ & $\checkmark$ & $\times$ \\
SpatialLM~\citeyearpar{mao2025spatialLM}          & Synthetic  & 12,328  & $\checkmark$ & $\checkmark$ & $\times$ \\
BIMNet~\citeyearpar{liu2025bimnet}                & Real-world & 25      & $\times$     & $\times$     & $\checkmark$ \\
CrabSet                                           & Synthetic  & 6,500   & $\times$     & $\times$     & $\checkmark$ \\
\bottomrule
\end{tabular}
}
\end{table}

%% file: figs/fig_dataset.tex
\begin{figure}[t]
  \centering
  \includegraphics[width=\linewidth]{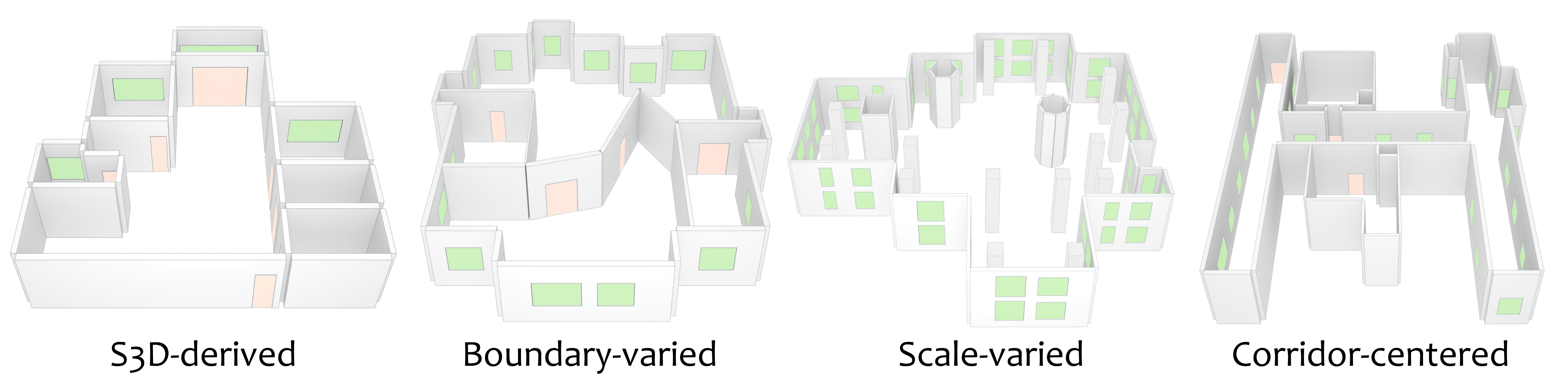}

\caption{
Representative samples from \emph{CrabSet}, combining S3D-derived residential and procedurally synthesized scenes for cuboid-level supervision.
The synthesized subsets vary boundaries, scales, and corridor-centered layouts, covering non-uniform wall thicknesses, non-Manhattan contours, service shafts, and diverse circulation patterns.
}

  \label{fig:dataset}
\end{figure}


%% file: src/04-method.tex
\begin{figure*}
  \centering
  \includegraphics[width=\linewidth]{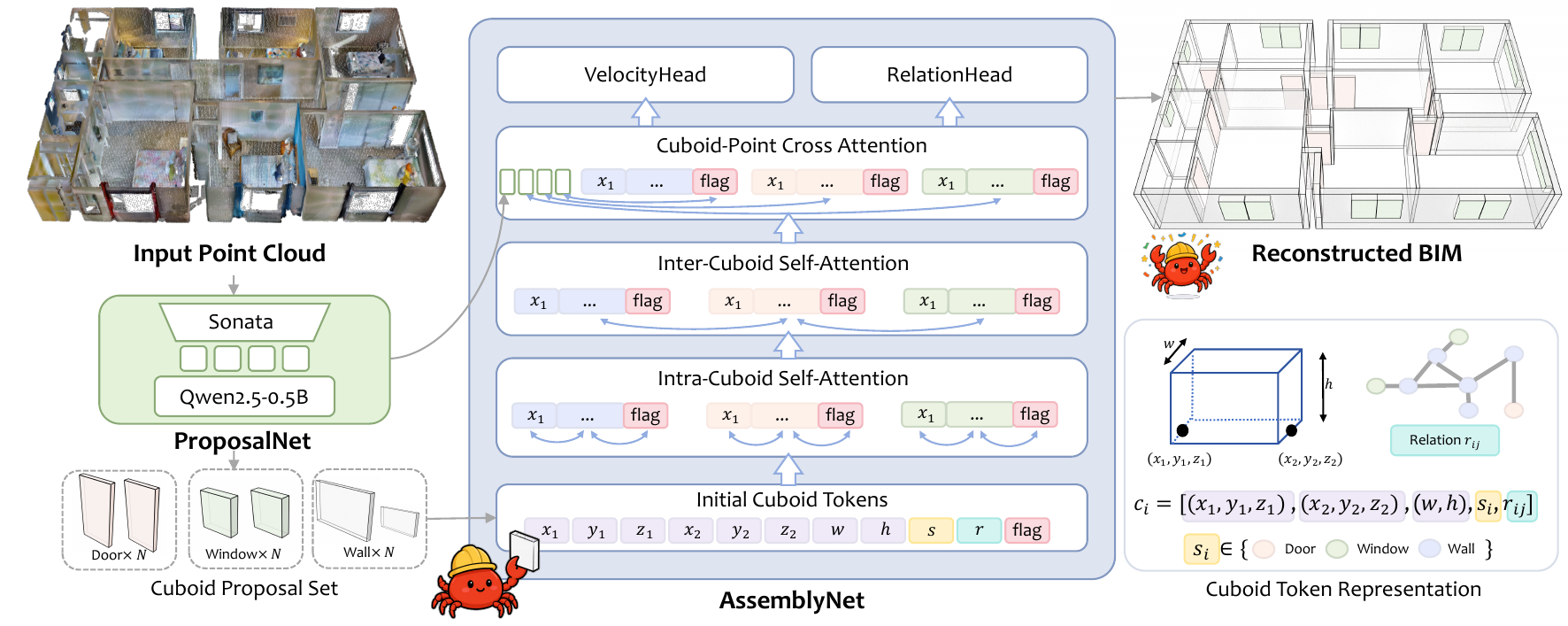}
\caption{{
Overview of \textbf{CRAB}. 
Given a raw indoor point cloud, \textit{ProposalNet} encodes the scan and autoregressively predicts an initial set of semantic cuboids for walls, doors, and windows. 
\textit{AssemblyNet} represents each proposal with centerline-based cuboid tokens augmented by semantic, relation, and edit-control flags, and refines them through intra-cuboid, inter-cuboid, and cuboid--point attention. 
A velocity head performs relation-aware flow-matching refinement of cuboid geometry, while a relation head infers structural dependencies such as wall adjacency and door/window affiliation. 
The refined cuboids and relations are finally instantiated as editable BIM-ready components, converting unstructured scans into component-level architectural models.
}}
  \label{fig:network}
\end{figure*}

\section{Methodology}\label{sec:method}

\subsection{Problem Formulation}

\hr{Our goal is to reconstruct as-built BIM components, such as walls, doors, windows, beams, and columns, from an input point cloud. We represent them as parameterized semantic cuboids along with their explicit topological relationships. }

Formally, we define the input point cloud as \(P=\{p_i\}\), where each point feature $p_i$ contains spatial coordinates and color information. The reconstructed architectural components are represented by a set of semantic cuboids
$$C=\{c_i\}, \quad c_i=[g_i,s_i], \quad g_i=[x_1,y_1,z_1,x_2,y_2,z_2,w,h],$$
where the geometric parameters $g_i$ define the bottom central axis connecting endpoints $(x_1,y_1,z_1)$ and $(x_2,y_2,z_2)$, together with cuboid thickness $w$ and height $h$, while $s_i$ denotes the semantic category. 

The relationships between cuboids are further represented by:
$$R=\{r_{ij} \mid c_i, c_j \in C\},$$
where $r_{ij}$ encodes structural interactions between components, including attachment relationships (\eg, doors or windows belonging to a wall) and adjacency relationships between neighboring walls.

The overall reconstruction task is therefore formulated as
$$(C, R) = \mathcal{F}(P),$$
where $\mathcal{F}$ denotes the proposed mapping function that reconstructs cuboids and their relationships from the input point cloud.

\subsection{ProposalNet}

We argue that the primary objective of the first stage is to generate accurate individual architectural components, since reliable prediction of each cuboid forms the foundation for subsequent relationship reasoning and structural assembly. Therefore, this module focuses exclusively on cuboid geometry and semantic prediction. Formally, we aim to learn the conditional distribution $p_\theta(C \mid P)$, from which the initial cuboid proposals are generated:
$$C_{\text{init}} \sim p_\theta(C \mid P).$$

Specifically, we employ \textit{Sonata}~\cite{wu2025sonata} as the point cloud encoder and \textit{Qwen2.5-0.5B}~\cite{an2024qwen} as the auto-regressive cuboid generator. 
The Sonata encoder is initialized with pretrained weights and further trained on our point cloud and cuboid pairs. 
For each input scene, we voxel-downsample the point cloud at \(2.5\,\mathrm{cm}\) and randomly keep at most \(800{,}000\) points for training and inference.

The point cloud encoder and autoregressive cuboid decoder are connected by a lightweight MLP Adapter, which maps point cloud features into the decoder token embedding space. 
Conditioned on the adapted point cloud tokens, the decoder sequentially generates the initial semantic cuboids. 
We fine-tune the generator on cuboid token sequences, with a maximum of 200 cuboids and an EOS token for termination. 
The proposal process is formulated as:

$$C_{\text{init}} = \text{AR}_\theta \Big( \text{Adapter}\big(\text{Enc}(P)\big) \Big),$$

where $\text{Enc}(\cdot)$ denotes the point cloud encoder, 
$\text{Adapter}(\cdot)$ maps point cloud features into the decoder token embedding space, 
and $\text{AR}_\theta$ represents the autoregressive transformer that decodes the adapted point cloud tokens into initial cuboid proposals.

\subsection{AssemblyNet}

The cuboid proposal stage produces an initial set of architectural elements, but the predicted cuboids may still contain geometric inaccuracies and structural inconsistencies. AssemblyNet aims to jointly refine cuboid geometry and infer structural relationships between components through Flow Matching. It also supports effective user-guided local editing. Formally, this stage is defined as
\[
(C_{\mathrm{revised}}, R) = \mathcal{A}_{\phi}(P, C_{\mathrm{init}}),
\]
where $\mathcal{A}_{\phi}$ denotes the AssemblyNet, $C_{\mathrm{revised}}$ is the refined cuboids.

Specifically, for each cuboid, the input to \emph{AssemblyNet} consists of explicit cuboid parameter tokens, one relation-aware token, and a change flag, together with point cloud features used for cross attention. 
The parameter tokens encode the cuboid geometry and semantic type, while the relation-aware token captures the structural role of this cuboid and participates in all attention operations.

The tokens are first embedded with parameter-specific linear layer encoders and timestep embeddings. They are then processed by an Intra-Cuboid Attention module, which models interactions among tokens within each cuboid and helps distinguish fixed cuboids from refinement targets according to the change flag. Next, Inter-Cuboid Attention propagates information across cuboids, allowing each cuboid to perceive the positions, scales, semantic categories, and potential structural dependencies of neighboring cuboids. A Cuboid-Point Cross Attention module further associates cuboid tokens with point cloud features, grounding the refinement process in the observed scene geometry. Details of the three attention modules are provided in the supplementary material.

Inspired by RectifiedFlow editing~\cite{kulikov2025flowedit,gong2025instantedit}, we formulate cuboid assembly as an iterative refinement process from the coarse proposal \(C_{\mathrm{init}}\). 
The geometry-aware token features are decoded by an MLP-based velocity head to predict geometric updates, and the final refined cuboids are obtained by accumulating the updates over \(K\) steps:
\[
C_{\mathrm{revised}} = C^K
=
C_{\mathrm{init}} + \sum_{k=0}^{K-1}
\mathrm{VelocityHead}(Z^k)\,\Delta t,
\]
where \(K\) is the number of refinement steps, \(\Delta t=1/K\), and \(Z^k=\mathrm{Enc}_{\phi}(C^k,t_k,P,flag)\) denotes the geometry-aware token features at the \(k\)-th step. 
Here, \(C^k\) is the intermediate cuboid state, \(t_k\) is the timestep, \(P\) is the input point cloud, and \(flag\) indicates whether each cuboid is fixed or editable. 
This iterative process refines editable cuboids while preserving fixed components.

In parallel, the relation-aware tokens encoded by the preceding attention modules are decoded by \(\mathrm{RelationHead}\) to predict pairwise structural relationships \(R=\{r_{ij}\}\). 
For each cuboid pair, \(\mathrm{RelationHead}\) uses an order-invariant pair feature:
\[
r_{ij}
=
\mathrm{RelationHead}
\left(
[q_i^{r}+q_j^{r},\; |q_i^{r}-q_j^{r}|,\; q_i^{r}\odot q_j^{r}]
\right),
\]
where \(q_i^{r}\) and \(q_j^{r}\) are the encoded relation tokens of the two cuboids, and \(\odot\) denotes element-wise multiplication. 
This formulation produces an undirected graph that captures door-wall affiliation and wall-wall adjacency without hand-crafted geometric rules at inference. Training details are provided in the supplementary material.

\subsection{Reconstruction Pipeline} 
\label{sec:reconstruction_pipeline}

CRAB converts an input point cloud into an editable as-built BIM model through the following three-step pipeline.

\paragraph{Step 1: Cuboid Proposal.}
\hr{ProposalNet reconstructs most architectural components from the input
point cloud as an initial set of semantic cuboids with coarse geometric
parameters.}

\paragraph{Step 2: Human-in-the-Loop Assembly.}
\hr{User interaction is optional: users may add, delete, move, or anchor coarse
cuboids before assembly. AssemblyNet jointly uses the cuboid set and input
point cloud to refine component geometry and predict their relationships,
automatically correcting geometric errors, local misalignments, and redundant
components. Completely missing components require coarse user initialization,
after which AssemblyNet aligns and integrates them into the existing structure.}

\paragraph{Step 3: BIM Conversion.}
\hr{Refined cuboids and relations are converted into editable IFC components.
For walls, adjacency relations resolve junctions by trimming or extending
each wall only along its longitudinal axis, while preserving its direction,
thickness, and centerline offset. The resulting wall-adjacency graph is
searched for closed cycles containing at least four wall segments, which
are assembled into rooms in a bottom-up manner and can be further converted
into 2D floor plans or 3D planar representations. Opening-wall affiliations
associate doors and windows with their host walls for wall-consistent
placement. Based on the refined cuboids and predicted relations, the BIM
model can be instantiated with corner-joint, cross-joint, or wall-integrated
wall structures.}

\input{tab/tab_compare_sota}

%% file: tab/tab_compare_sota.tex
\begin{table*}[!t]
      \centering
      \scriptsize
      \belowrulesep=0pt
      \aboverulesep=0pt
      \setlength{\tabcolsep}{4pt}
      \renewcommand{\arraystretch}{1.12}
      \caption{
Quantitative comparison of representation choices on Structured3D.
\emph{CRAB} improves both component recovery and BIM-critical volumetric attributes, especially wall thickness and opening height, over floor plan and plane-based alternatives.
Best and second-best results are highlighted in green and blue.
}
      \label{tab:main_compare}
      \resizebox{\textwidth}{!}{
      \begin{tabular}{l|c|c|ccccc|cc|cc}
      \toprule
      \multirow{2}{*}{Method} & \multirow{2}{*}{Rep.}
      & \multicolumn{1}{c|}{Exterior Wall}
      & \multicolumn{5}{c|}{Interior Wall}
      & \multicolumn{2}{c|}{Door}
      & \multicolumn{2}{c}{Window} \\
      \cmidrule(lr){3-3}
      \cmidrule(lr){4-8}
      \cmidrule(lr){9-10}
      \cmidrule(lr){11-12}
      &
      & Match-F1 ($\uparrow$)
      & Match-F1 ($\uparrow$)
      & CL-Offset ($\downarrow$)
      & T-MAE ($\downarrow$)
      & T-Acc@3 ($\uparrow$)
      & T-Acc@5 ($\uparrow$)
      & Match-F1 ($\uparrow$)
      & H-MAE ($\downarrow$)
      & Match-F1 ($\uparrow$)
      & H-MAE ($\downarrow$) \\
      \midrule
      \hr{Cloud2BIM}~\citeyearpar{zbirovsky2025cloud2bim} & Volumetric & 4.41\% & 62.76\% & -- & -- & -- & -- & 15.12\% & -- & 4.52\% & -- \\
      RoomFormer~\citeyearpar{yue2023roomformer}   & Floor Plan & 84.48\%            & 85.78\%            & 5.34              & 6.19              & 27.79\%             & 48.54\% & 77.05\%             & 74.94             & 64.82\%             & 120.84            \\
      CAGE~\citeyearpar{liu2025cage}         & Floor Plan & 91.14\%            & 89.74\%            & 3.68              & 5.31              & 33.75\%             & 58.11\% & --                  & --                & --                  & --                \\
      SceneScript$^{*}$  & Plane     & 44.40\%       & 52.59\%            & 6.02              & 9.24              & 10.73\%             & 16.26\% & 47.83\%             & 22.64             & 38.71\%             & 22.01             \\
      SpatialLM~\citeyearpar{mao2025spatialLM} & Plane     & 92.40\%            & 91.48\%            & 3.10              & 3.89              & 24.98\%             & 77.74\% & 89.03\%             & 77.36             & 77.90\%             & 122.86            \\
      SpatialLM$^{*}$ & Plane     & 91.52\%       & 91.03\%            & \seconddown{3.09} & 4.38              & 41.62\%             & 70.60\% & 81.44\%             & 78.72             & 65.67\%             & 78.26             \\
      Ours$^{*}$ \hr{(ProposalNet)}    & Cuboid    & \secondup{92.81\%} & \secondup{91.91\%} & 3.35              & \seconddown{2.95} & \secondup{79.15\%}  & \secondup{86.18\%}  & \secondup{88.07\%}  & \seconddown{3.63} & \secondup{86.52\%}  & \seconddown{9.52} \\
      Ours \hr{(ProposalNet)}    & Cuboid    & \bestup{93.78\%}   & \bestup{92.57\%}   & \bestdown{3.01}   & \bestdown{2.80}   & \bestup{79.50\%}    & \bestup{87.17\%}    & \bestup{91.14\%}    & \bestdown{2.86}   & \bestup{88.97\%}    & \bestdown{9.23}   \\
      \bottomrule
      \end{tabular}
      }
      \vspace{2pt}

\begin{flushleft}
\footnotesize
\textbf{Note.}
All geometric errors are reported in centimeters.
$^{*}$ denotes models retrained on Structured3D with representation-specific annotations and ground-truth opening heights.
Unmarked baselines use officially released weights.
Ours is trained on the full \emph{CrabSet}, while Ours$^{*}$ uses only S3D-derived cuboid annotations for controlled comparison.
\end{flushleft}
\end{table*}

%% file: src/05-results.tex
\section{Results and Evaluation}

\subsection{Experimental Setup}

\paragraph{Datasets.}

For training, we use 3,000 Structured3D training scenes and 3,000 procedurally generated scenes from \emph{CrabSet}. 
For evaluation, we evaluate on the
Structured3D validation and test splits (250 scenes each), with BIMNet
and ISPRS for real-scene generalization.

\paragraph{Metrics.} We evaluate reconstruction quality in two stages. \textbf{Match-F1} measures coarse component-level matching quality using the F1 score computed from precision and recall based on voxel-overlap matching. 
After coarse matching, geometric metrics evaluate accuracy: \textbf{CL-Offset} reports the centerline deviation of interior walls in centimeters, \textbf{T-MAE} measures the mean thickness error of walls, and \textbf{T-Acc@}$k$ denotes the proportion of matched walls whose thickness error is within $k$\, cm. \textbf{H-MAE} measures height errors for doors and windows. \hr{Details} are provided in the supplementary material.

\paragraph{Baselines.}
We compare \emph{CRAB} with \hr{an open-source volumetric Scan-to-BIM method, Cloud2BIM\cite{zbirovsky2025cloud2bim},} floorplan-based methods, including RoomFormer\cite{yue2023roomformer} and CAGE\cite{liu2025cage}, and 3D plane-based methods, including SceneScript\cite{avetisyan2024scenescript} and SpatialLM\cite{mao2025spatialLM}. 
Since released plane-based models lack valid opening heights on Structured3D, we report SceneScript$^{*}$ and SpatialLM$^{*}$, fine-tuned on corrected S3D annotations. 
We report Ours$^{*}$, trained only on S3D-derived cuboids, and Ours, trained on \emph{CrabSet}, to ensure fair comparison and assess procedural cuboid supervision.

\paragraph{Training Details.} The model is trained for 50 epochs using the Adam optimizer with an initial learning rate of $1\times10^{-4}$. The learning rate follows a cosine decay schedule with a 1\% warm-up phase. All experiments were conducted on a server equipped with 8 NVIDIA GeForce RTX 4090 GPUs and dual Intel Xeon Gold 6430 CPUs.

\paragraph{Geometric Granularity and Post-processing.}
For fair component-level evaluation, we follow the original discretization settings of each baseline whenever possible and convert floor plan or plane outputs into volumetric component representations before evaluation. 
Since these baselines do not directly output cuboid-level walls, doors, and windows, additional conversion is required before evaluation. 
Detailed discretization settings and post-processing procedures are provided in the supplementary material.

\subsection{Evaluation}

We evaluate \emph{CRAB} from six perspectives: 
\emph{(i) volumetric accuracy} on Structured3D, covering component matching, wall thickness, and opening height; 
\hr{\emph{(ii) layout and plane quality} using the native metrics of representative 2D floor plan and 3D plane methods;}
\emph{(iii) robustness} under structural ambiguity and local perturbations; 
\emph{(iv) real-world generalization} on noisy and incomplete BIMNet and ISPRS scans; 
\emph{(v) editability and BIM conversion} enabled by cuboid components and their relations; and 
\hr{\emph{(vi) beam and column reconstruction}  on procedurally generated scenes to assess broader coverage.}

\paragraph{Volumetric Reconstruction Accuracy.}
Cuboid representations directly model architectural elements as 3D components, avoiding the need to infer wall thickness or opening geometry from 2D layouts or planar fragments. 
We evaluate these BIM-critical attributes on the Structured3D test set in Table~\ref{tab:main_compare}. 
\hr{Compared with the volumetric baseline Cloud2BIM, \emph{CRAB}
produces more complete component reconstructions on incomplete scans, with
fewer fragmented walls and unstable openings.}
Compared with floor-plan and plane-based alternatives, \emph{CRAB} achieves stronger component matching and more accurate geometry. 
For openings, \emph{CRAB} improves window Match-F1 from 77.90\% to 88.97\% over SpatialLM, and reduces height errors to 2.86\,cm for doors and 9.23\,cm for windows. 
For interior walls, \emph{CRAB} achieves a wall-thickness MAE of 2.80\,cm and a T-Acc@3 of 79.50\%, substantially outperforming the strongest non-cuboid baseline at 41.62\% T-Acc@3. 
As shown in Fig.~\ref{fig:compare_thickness}, \emph{CRAB} also better preserves wall thickness, wall extent, and opening--wall consistency.

\input{figs/fig_compare_thickness} 

\paragraph{Layout and Plane Quality.} \hr{As shown in Table~\ref{tab:layout_plane_quality}, we evaluate CRAB using RoomFormer's native 2D layout metrics~\cite{yue2023roomformer} and SpatialLM's native 3D plane metrics~\cite{mao2025spatialLM}. Following Sec.~\ref{sec:reconstruction_pipeline}, CRAB outputs are converted into 2D room polygons and 3D planes for evaluation. For cross-representation comparison, plane-based outputs are grouped into rooms by matching shared wall endpoints, while floorplan outputs are lifted using point-cloud heights. CRAB achieves notably higher Corner F1 and Angle F1 than CAGE, demonstrating more accurate boundary geometry. Compared with SpatialLM, CRAB achieves comparable wall-plane performance and improves window F1 from 87.5\% to 91.4\% at IoU$_{2D}$@0.5. Overall, CRAB achieves competitive layout and plane quality against baselines while providing accurate volumetric reconstruction.}

\input{tab/tab_layout_plane_compare} 

\input{figs/fig_assembly_cell_test} 

\paragraph{Robustness to Ambiguity and Local Perturbations.}
Cuboids provide more reliable refinement units because each prediction corresponds to an explicit volumetric component. 
This is particularly useful in incomplete, ambiguous, or non-Manhattan scenes, where floor-plan methods depend on room closure and plane-based methods may become unstable when several nearby planes describe the same wall or opening. 
As shown in Fig.~\ref{fig:compare_sota}, \emph{CRAB} recovers more coherent wall structures and more consistent opening geometry by directly predicting component-level cuboids.

We further test robustness with a controlled wall-perturbation experiment. 
A ground-truth wall is locally shifted or resized to simulate imperfect editing, and the model is asked to refine it back to the original geometry. 
\hr{Plane-based refinement produces endpoint errors of 30.29\,cm, 13.07\,cm, and 33.21\,cm under length-plus-axial, perpendicular, and combined perturbations, respectively, while cuboid-based refinement reduces them to 3.60\,cm, 5.18\,cm, and 5.66\,cm. As shown in Fig.~\ref{fig:assembly_cell_test}, } 
plane-based refinement can drift toward nearby planar structures, whereas cuboid-based refinement preserves the target wall as a distinct volumetric instance.

We also evaluate full-scene refinement by making all predicted components editable. 
\hr{Specifically, we use SpatialLM's plane predictions and CRAB's cuboid predictions reported in Table~\ref{tab:main_compare} as inputs to the plane-based and cuboid-based variants of AssemblyNet, respectively.}
Table~\ref{tab:assembly_proposal_test} shows that plane-based refinement can degrade the original prediction, since small plane deviations are amplified during post-processing for wall extraction and thickness estimation. 
In contrast, cuboid-based AssemblyNet preserves explicit component geometry and improves interior-wall T-Acc@3 from 79.50\% to 81.27\%. 
These results indicate that cuboids are more stable refinement targets for Scan-to-BIM reconstruction, as each editable element already carries its own extent and thickness parameters.

\input{tab/tab_assembly_proposal_test} 

\paragraph{Real-World Generalization.}
We compare \emph{CRAB} with SpatialLM on real scenes to evaluate generalization beyond synthetic scenes. 
\hr{Since real-scene benchmarks lack consistent cuboid annotations, we
evaluate BIMNet using Match-F1 computed from voxel overlap. CRAB achieves 72.5\%, 68.8\%, and 48.5\% for walls, doors, and windows, compared with 68.6\%, 49.6\%, and 37.4\% for SpatialLM. The larger gains on openings indicate more reliable recovery of BIM-ready components.}
As shown in Fig.~\ref{fig:compare_real}, real scenes contain complex layouts, sensor noise, missing regions, and domain gaps from synthetic scenes. 
Under these challenging conditions, SpatialLM often produces incomplete or unstable structures. 
In contrast, \emph{CRAB} recovers more coherent layouts and more accurate components. 
This trend is observed on both residential-like BIMNet scenes and non-residential ISPRS scenes, indicating that the cuboid representation learned from \emph{CrabSet} generalizes better to diverse real-world scans.

\paragraph{Editability and BIM Conversion.}
Cuboid representation also provides practical flexibility for editing and BIM conversion. 
Unlike floor plan or plane-based outputs, where wall and opening edits must be inferred through polygon updates, plane grouping, or post-processing, \emph{CRAB} represents architectural components as independent volumetric instances. 
Thus, users can directly insert, delete, translate, or resize component instances in a BIM-like manner.
As shown in Fig.~\ref{fig:bim_conversion}, the predicted cuboids and relations further support flexible BIM instantiation. 
The same cuboid-relation graph can be converted into different modeling styles, including \textit{Corner-Joint BIM}, \textit{Cross-Joint BIM}, and \textit{Wall-Integrated BIM}. 
This demonstrates that \emph{CRAB} produces not only accurate geometry, but also editable and relation-aware components for downstream architectural modeling.

\paragraph{Beam and Column Reconstruction.}
\hr{Beyond walls, doors, and windows, CRAB also supports beams and columns. Since existing benchmarks lack consistent annotations and reproducible baselines for these categories, we evaluate them separately on procedurally generated scenes. CRAB achieves Match-F1 scores of 97.15\% for columns and 57.83\% for beams. See supplementary material for details.}

\input{figs/fig_bim_convertion}  

\subsection{Ablation Study}

\paragraph{Effect of Centerline-based Parameterization.}
We ablate the proposed centerline-based cuboid parameterization against a generic oriented box representation, which can be viewed as a simplified variant of common primitive and scene tokenizations~\cite{ye2025primitiveanything,feng2025casagpt}. 
Since architectural components are vertically extruded in our setting, the oriented box is represented by center translation, side lengths, and a yaw rotation angle around the vertical axis. 
Both variants are trained under the same setting, with only the geometric representation changed.

As shown in Table~\ref{tab:ablation_cuboid}, oriented boxes perform slightly better on doors and achieve comparable results on windows, which are relatively compact components. 
However, for long and thin walls, the centerline-based cuboid explicitly separates wall extent from wall thickness, leading to better wall reconstruction and improving interior-wall T-Acc@3 from 71.74\% to 79.50\%. 
For highly elongated architectural components, longitudinal extent and thickness play strongly asymmetric geometric roles, and centerline-based parameterization better disentangles them, reducing geometric ambiguity and improving wall reconstruction stability.
This suggests that the proposed parameterization is more suitable for wall-dominated architectural reconstruction, where accurate wall geometry is critical for downstream BIM conversion.

\subsection{Limitation Analysis}

\emph{CRAB} still has several limitations. 
\hr{\emph{(1) Missing component recovery.}
\emph{CRAB} can automatically correct continuous geometric errors, local misalignments, and redundant components, but entirely missing components may still require coarse user initialization. The supplementary material further shows \emph{AssemblyNet}'s ability to correct geometric errors and local misalignments, while removing redundant components through component alignment.}
\emph{(2) Irregular non-Manhattan geometry.} 
While \emph{CRAB} handles standard Manhattan layouts and moderate non-Manhattan variations, highly irregular or curved structures remain challenging: cuboids can capture the dominant organization, but may oversimplify fine-scale deviations. 
\emph{(3) Sparse infrequent components.} 
Beams, columns, stairs, and service shafts appear sparsely in current real-world test scenes, limiting comprehensive evaluation. 
Although \emph{CRAB} can model these elements in principle, robust reconstruction requires richer component-level supervision and broader BIM benchmarks.

\input{tab/tab_ablation_cuboid}  

%% file: figs/fig_compare_thickness.tex
\begin{figure}[t]
  \centering
  \includegraphics[width=\linewidth]{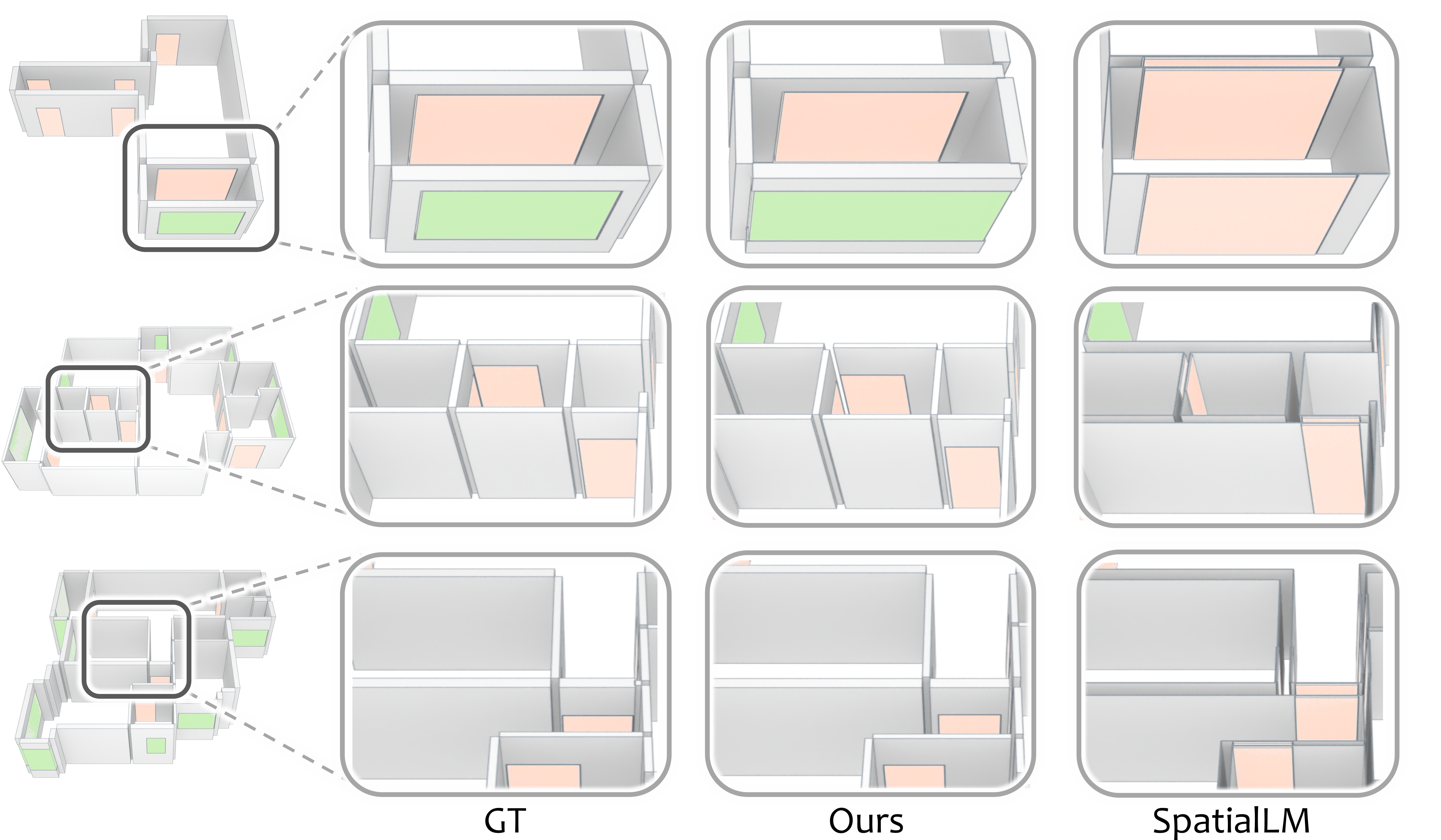}
  \caption{Qualitative comparison with a plane-based representation.
Zoomed views compare the same local structures in GT, \emph{CRAB}, and SpatialLM.
\emph{CRAB}'s volumetric cuboids better preserve wall thickness, component extent, and opening-wall consistency than the plane-based representation.}
  \label{fig:compare_thickness}
\end{figure}

%% file: tab/tab_layout_plane_compare.tex
\begin{table*}[!t]
    \centering
    \scriptsize
    \belowrulesep=0pt
    \aboverulesep=0pt
    \setlength{\tabcolsep}{2.6pt}
    \renewcommand{\arraystretch}{1.12}
    \caption{
    \hr{Quantitative comparison of 2D layout and 3D plane quality on Structured3D. The results show that CRAB’s gains in volumetric reconstruction do not come at the expense of layout or plane quality. Best and second-best results are highlighted in green and blue.}
    }
    \label{tab:layout_plane_quality}

    \resizebox{\textwidth}{!}{
    \begin{tabular}{l|c|ccc|ccc|ccc|cccc|cccc}
    \toprule
    \multirow{3}{*}{Method}
    & \multirow{3}{*}{Rep.}
    & \multicolumn{9}{c|}{2D Layout}
    & \multicolumn{8}{c}{3D Plane} \\
    \cmidrule(lr){3-11}
    \cmidrule(lr){12-19}

    &
    & \multicolumn{3}{c|}{Room}
    & \multicolumn{3}{c|}{Corner}
    & \multicolumn{3}{c|}{Angle}
    & \multicolumn{4}{c|}{IoU$_{\mathrm{2D}}$@0.25}
    & \multicolumn{4}{c}{IoU$_{\mathrm{2D}}$@0.5} \\
    \cmidrule(lr){3-5}
    \cmidrule(lr){6-8}
    \cmidrule(lr){9-11}
    \cmidrule(lr){12-15}
    \cmidrule(lr){16-19}

    &
    & Prec. & Rec. & F1
    & Prec. & Rec. & F1
    & Prec. & Rec. & F1
    & Avg. & Wall & Door & Window
    & Avg. & Wall & Door & Window \\
    \midrule

    RoomFormer~\citeyearpar{yue2023roomformer}
    & Floor Plan
    & 98.7
    & 96.7
    & 97.3
    & 89.2
    & 85.3
    & 87.2
    & 83.0
    & 79.6
    & 81.3
    & 83.4
    & 86.0
    & 85.1
    & 78.9
    & 81.4
    & 84.6
    & 83.1
    & 76.6 \\

    CAGE~\citeyearpar{liu2025cage}
    & Floor Plan
    & \bestup{99.6}
    & \bestup{98.7}
    & \bestup{99.1}
    & \secondup{95.0}
    & 88.6
    & 91.7
    & 92.5
    & 86.4
    & \secondup{89.3}
    & --
    & 89.3
    & --
    & --
    & --
    & 86.4
    & --
    & -- \\

    SceneScript
    & Plane
    & 81.3
    & 45.3
    & 58.2
    & 71.1
    & 29.8
    & 42.0
    & 68.4
    & 28.6
    & 40.4
    & 90.4
    & 92.9
    & 92.8
    & 85.5
    & 89.2
    & 91.9
    & 92.5
    & 83.3 \\

    SpatialLM~\citeyearpar{mao2025spatialLM}
    & Plane
    & \secondup{99.3}
    & \secondup{97.2}
    & 98.2
    & 94.5
    & \secondup{89.5}
    & \secondup{92.0}
    & \bestup{93.7}
    & \secondup{88.7}
    & \bestup{91.1}
    & \secondup{92.2}
    & \bestup{93.5}
    & \secondup{94.1}
    & \secondup{88.9}
    & \secondup{91.1}
    & \bestup{92.4}
    & \secondup{93.4}
    & \secondup{87.5} \\

    CRAB (Ours)
    & Cuboid
    & 98.7
    & \bestup{98.7}
    & \secondup{98.7}
    & \bestup{95.5}
    & \bestup{90.7}
    & \bestup{93.1}
    & \secondup{93.4}
    & \bestup{88.8}
    & \bestup{91.1}
    & \bestup{93.3}
    & \secondup{93.2}
    & \bestup{94.6}
    & \bestup{92.1}
    & \bestup{92.6}
    & \secondup{92.2}
    & \bestup{94.1}
    & \bestup{91.4} \\

    \bottomrule
    \end{tabular}
    }

    \vspace{2pt}
    \begin{flushleft}
    \footnotesize
    \end{flushleft}
\end{table*}

%% file: figs/fig_assembly_cell_test.tex
\begin{figure}[t]
  \centering
  \includegraphics[width=\linewidth]{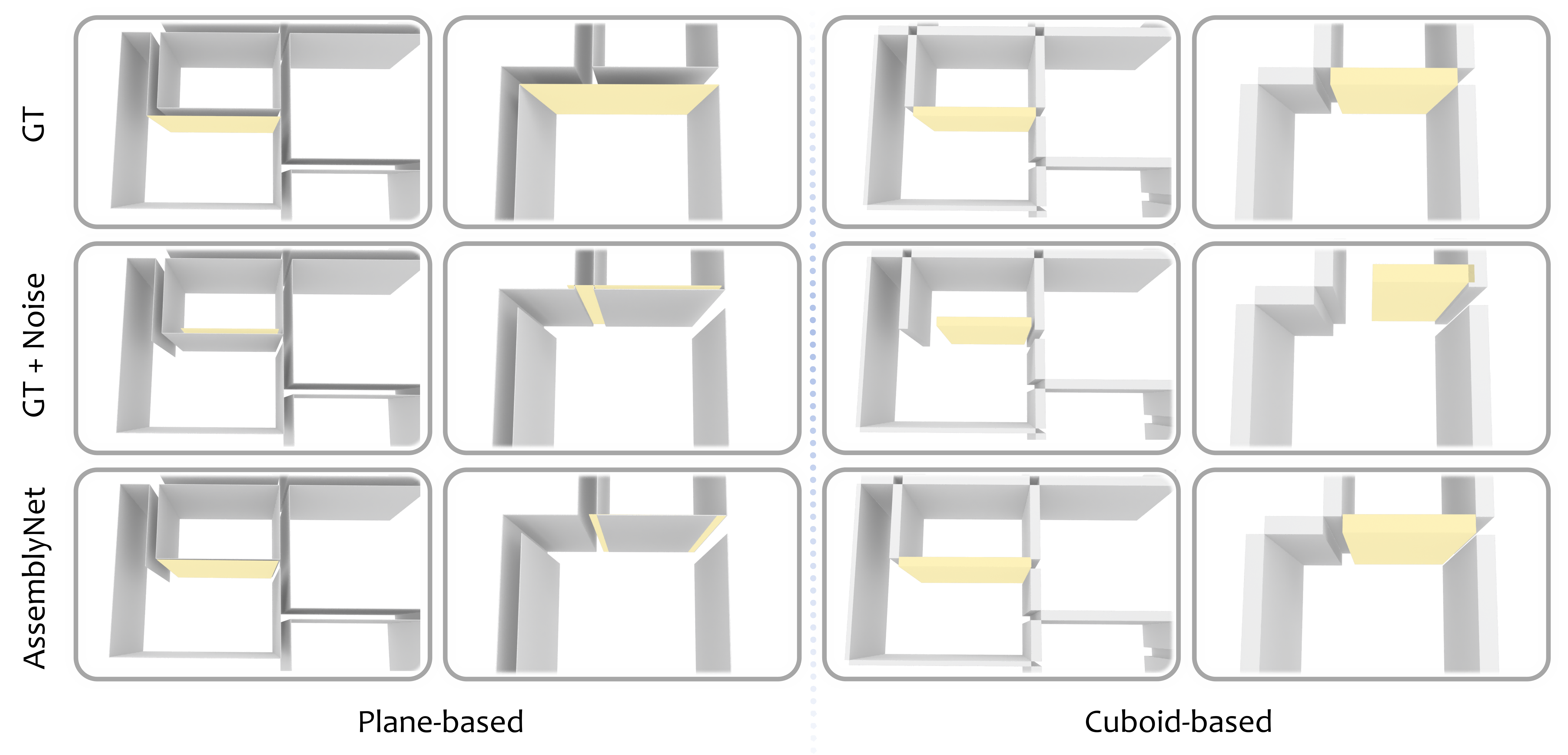}
  \caption{
  \emph{AssemblyNet} refinement under simulated wall editing errors. 
The three rows show the ground truth, the perturbed input, and the refined result, respectively.
Starting from the noisy wall highlighted in yellow, plane-based refinement can drift toward nearby structures, while cuboid-based refinement better preserves and recovers the target volumetric component.}
  \label{fig:assembly_cell_test}
\end{figure}

%% file: tab/tab_assembly_proposal_test.tex




\begin{table}[t]
\centering
\scriptsize
\belowrulesep=0pt
\aboverulesep=0pt
\setlength{\tabcolsep}{3.5pt}
\renewcommand{\arraystretch}{1.12}
\caption{\hr{Comparison of full-scene refinement with plane-based and cuboid-based AssemblyNet, using planes from SpatialLM and cuboids from ProposalNet, respectively. Cuboid-based AssemblyNet improves the initial reconstruction results, whereas plane-based AssemblyNet degrades them.}}
\label{tab:assembly_proposal_test}
\resizebox{\linewidth}{!}{
\begin{tabular}{cccccc}
\toprule
\multirow{2}{*}{Method}
& \multicolumn{1}{c}{Exterior Wall}
& \multicolumn{2}{c}{Interior Wall}
& \multicolumn{1}{c}{Door}
& \multicolumn{1}{c}{Window} \\
\cmidrule(lr){2-2}\cmidrule(lr){3-4}\cmidrule(lr){5-5}\cmidrule(lr){6-6}
& Match-F1 ($\uparrow$) & Match-F1 ($\uparrow$) & T-Acc@3 ($\uparrow$) & Match-F1 ($\uparrow$) & Match-F1 ($\uparrow$) \\
\midrule
AssemblyNet (Plane)  & 87.94\% & 87.31\% & 20.45\% & 88.02\% & 76.44\% \\
AssemblyNet (Cuboid) & \textbf{94.42\%} & \textbf{92.73\%} & \textbf{81.27\%} & \textbf{92.20\%} & \textbf{89.76\%} \\
\bottomrule
\end{tabular}
}
\end{table}

%% file: figs/fig_bim_convertion.tex
\begin{figure}[t]
  \centering
  \includegraphics[width=\linewidth]{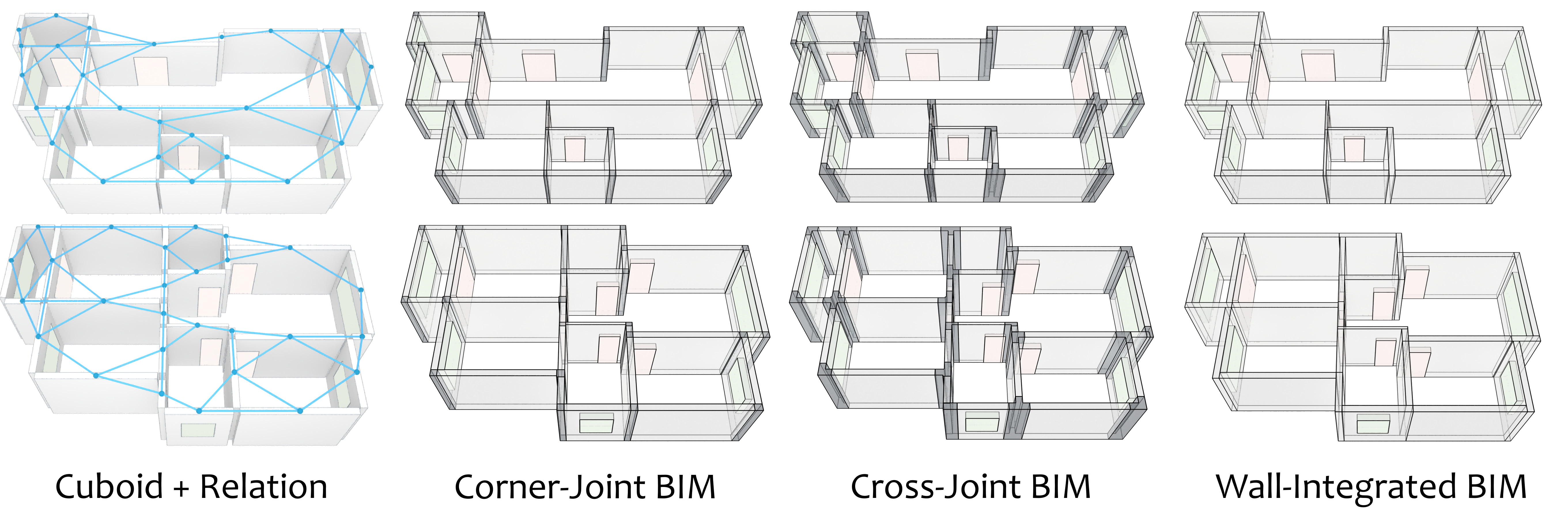}
  \caption{BIM conversion from cuboids and relations. 
The same cuboid-relation graph supports different junction assignments: \textit{Corner-Joint BIM} uses explicit corner joints, \textit{Cross-Joint BIM} uses explicit cross joints, and \textit{Wall-Integrated BIM} merges junction regions into connected wall elements. }
  \label{fig:bim_conversion}
\end{figure}

%% file: tab/tab_ablation_cuboid.tex
\begin{table}[t]
\centering
\scriptsize
\belowrulesep=0pt
\aboverulesep=0pt
\setlength{\tabcolsep}{3.5pt}
\renewcommand{\arraystretch}{1.12}
\caption{Ablation of cuboid parameterization.
Oriented boxes perform well on compact components such as doors and windows, but are less suitable for long and thin walls, where longitudinal extent and thickness play strongly asymmetric geometric roles.
Our centerline-based parameterization explicitly separates these two factors, improving wall geometry and thickness reconstruction and making it more suitable for BIM-oriented reconstruction.}
\label{tab:ablation_cuboid}
\resizebox{\linewidth}{!}{
\begin{tabular}{cccccc}
\toprule
\multirow{2}{*}{Method}
& \multicolumn{1}{c}{Exterior Wall}
& \multicolumn{2}{c}{Interior Wall}
& \multicolumn{1}{c}{Door}
& \multicolumn{1}{c}{Window} \\
\cmidrule(lr){2-2}\cmidrule(lr){3-4}\cmidrule(lr){5-5}\cmidrule(lr){6-6}
& Match-F1 ($\uparrow$) & Match-F1 ($\uparrow$) & T-Acc@3 ($\uparrow$) & Match-F1 ($\uparrow$) & Match-F1 ($\uparrow$) \\
\midrule
Oriented Box  & 92.59\% & 92.06\% & 71.74\% & \textbf{92.20\%} & 88.94\% \\
Centerline Cuboid (Ours)  & \textbf{93.78\%} & \textbf{92.57\%} & \textbf{79.50\%} & 91.14\% & \textbf{88.97\%} \\
\bottomrule
\end{tabular}
}
\end{table}

%% file: src/06-future.tex
\section{Conclusion and Future Work}

We presented \emph{CRAB}, a cuboid-based approach to indoor Scan-to-BIM reconstruction. 
The key observation is that BIM reconstruction requires component-level volumetric structure, not only layout curves or planar surfaces. 
Accordingly, \emph{CRAB} represents \hr{walls, doors, windows, beams and columns} as semantic cuboids with explicit centerline, thickness, height, and opening parameters. 
This representation makes architectural components directly editable and provides a natural interface for downstream BIM conversion.

The proposed framework follows a proposal-and-assembly design. 
\emph{ProposalNet} predicts coarse architectural cuboids from point-cloud observations, while \emph{AssemblyNet} refines their geometry and relationships through relation-aware Flow Matching. Together, the two-stage system is capable of correcting geometric errors and local misalignments while removing redundant components through component alignment.
To support learning and evaluation, we constructed \emph{CrabSet}, which provides cuboid-level supervision from Structured3D-derived annotations and procedurally generated architectural scenes. 
Experiments on synthetic and real-world scans show that cuboid assembly yields more accurate wall-thickness estimation, more stable refinement, and more flexible BIM instantiation than floor-plan- or plane-based alternatives. Importantly, these gains in volumetric reconstruction do not come at the expense of layout or plane quality, where \emph{CRAB} remains competitive with representation-specific baselines.

To address the limitations discussed earlier, future work will primarily broaden \emph{CRAB}'s practical capabilities and application scope. We will explore (1) complex non-Manhattan geometry and richer primitives, including cylinders and curved surfaces, (2) broader BIM categories and structural relations, together with semantic plane cues as auxiliary geometric evidence for robust reconstruction under real-scene domain shifts, and (3) interactive or agent-assisted workflows, reducing manual effort and enabling more automated Scan-to-BIM reconstruction.

%% file: figs/fig_compare_sota.tex
\begin{figure*}[t]
  \centering
  \includegraphics[width=\linewidth]{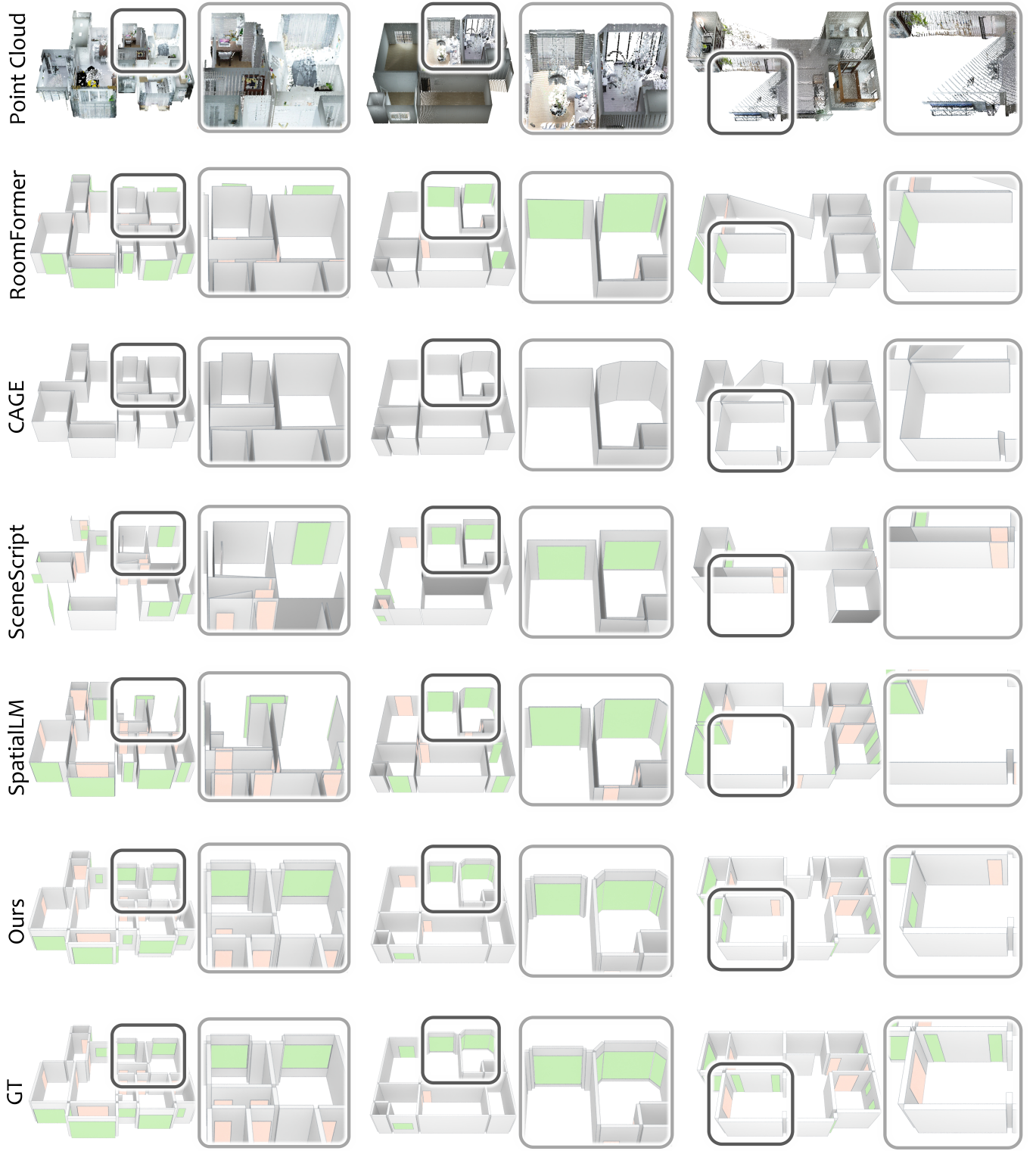}
  \caption{Qualitative comparison on Structured3D. 
Each column shows one scene, with rows showing the input point cloud, floor plan baselines, plane baselines, our cuboid reconstruction, and the ground truth. 
Compared with RoomFormer, CAGE, SceneScript, and SpatialLM, \emph{CRAB} recovers more coherent wall structures and more accurate door and window geometry in ambiguous, non-Manhattan, and incomplete regions.}
  \label{fig:compare_sota}
\end{figure*}

%% file: figs/fig_compare_real.tex
\begin{figure*}[t]
  \centering
  \includegraphics[width=0.97\linewidth]{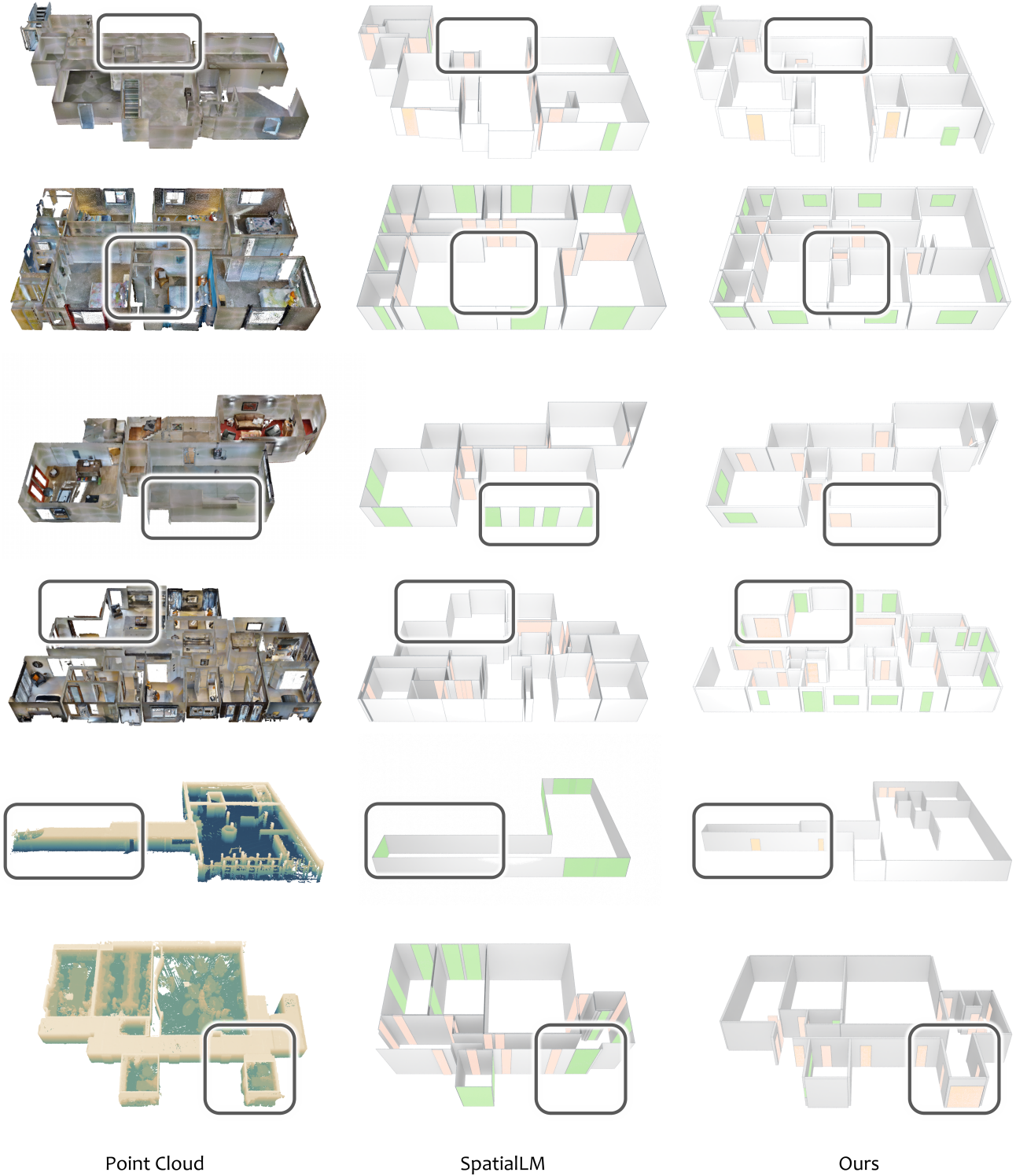}
  \caption{{Qualitative comparison on real-world scanned scenes. Each row shows one scene: input point cloud (left), SpatialLM (middle), and ours (right). \emph{CRAB} better preserves wall continuity and opening geometry under practical noise and clutter.}}
  \label{fig:compare_real}
\end{figure*}

%% file: CRAB.bib
@STRING{CVPR = {Proc. IEEE/CVF Conf. on Computer Vision \& Pattern Recognition}}

@STRING{ECCV = {Proc. Euro. Conf. on Computer Vision}}

@STRING{ICCV = {Proc. Int. Conf. on Computer Vision}}

@STRING{SIGGRAPH = {Proc. SIGGRAPH}}

@STRING{TOG = {ACM Trans. on Graphics}}

@STRING{TVCG = {IEEE Trans. Visualization \& Computer Graphics}}

@STRING{NIPS = {Proc. Conf. on Neural Information Processing Systems}}

@STRING{ISPRS = {ISPRS J. Photogrammetry and Remote Sensing}}

@inproceedings{yue2023roomformer,
  title={Connecting the dots: Floorplan reconstruction using two-level queries},
  author={Yue, Yuanwen and Kontogianni, Theodora and Schindler, Konrad and Engelmann, Francis},
  booktitle=CVPR,
  pages={845--854},
  year={2023}
}

@inproceedings{chen2023polydiffuse,
  title     = {PolyDiffuse: Polygonal Shape Reconstruction via Guided Set Diffusion Models},
  author    = {Chen, Jiacheng and Deng, Ruizhi and Furukawa, Yasutaka},
  booktitle = NIPS,
  volume    = {36},
  year      = {2023}
}

@inproceedings{avetisyan2024scenescript,
  title     = {SceneScript: Reconstructing Scenes With An Autoregressive Structured Language Model},
  author    = {Avetisyan, Armen and Xie, Christopher and Howard-Jenkins, Henry and Yang, Tsun-Yi and Aroudj, Samir and Patra, Suvam and Zhang, Fuyang and Frost, Duncan and Holland, Luke and Orme, Campbell and Engel, Jakob and Miller, Edward and Newcombe, Richard and Balntas, Vasileios},
  booktitle = ECCV,
  year      = {2024}
}

@inproceedings{mao2025spatialLM,
  title     = {SpatialLM: Training Large Language Models for Structured Indoor Modeling},
  author    = {Mao, Yongsen and Zhong, Junhao and Fang, Chuan and Zheng, Jia and Tang, Rui and Zhu, Hao and Tan, Ping and Zhou, Zihan},
  booktitle = NIPS,
  year      = {2025}
}

@article{chen2022pqtransformer,
  title   = {PQ-Transformer: Jointly Parsing 3D Objects and Layouts from Point Clouds},
  author  = {Chen, Xiaoxue and Zhao, Hao and Zhou, Guyue and Zhang, Ya-Qin},
  journal = {IEEE Robotics and Automation Letters},
  volume  = {7},
  number  = {2},
  pages   = {2519--2526},
  year    = {2022}
}

@inproceedings{chen2022heat,
  title     = {HEAT: Holistic Edge Attention Transformer for Structured Reconstruction},
  author    = {Chen, Jiacheng and Qian, Yiming and Furukawa, Yasutaka},
  booktitle = CVPR,
  pages     = {3866--3875},
  year      = {2022}
}

@inproceedings{chen2019floorsp,
  title     = {Floor-SP: Inverse CAD for Floorplans by Sequential Room-Wise Shortest Path},
  author    = {Chen, Jiacheng and Liu, Chen and Wu, Jiaye and Furukawa, Yasutaka},
  booktitle = CVPR,
  pages     = {2661--2670},
  year      = {2019}
}

@inproceedings{xu2024frinet,
  title     = {FRI-Net: Floorplan Reconstruction via Room-wise Implicit Representation},
  author    = {Xu, Honghao and Xu, Juzhan and Huang, Zeyu and Xu, Pengfei and Huang, Hui and Hu, Ruizhen},
  booktitle = ECCV,
  year      = {2024}
}

@article{liu2025bimnet,
  title   = {Dataset and Benchmark for As-built BIM Reconstruction from Real-world Point Cloud},
  author  = {Liu, Yudong and Huang, Han and Gao, Ge and Ke, Ziyi and Li, Shengtao and Gu, Ming},
  journal = {Automation in Construction},
  volume  = {173},
  pages   = {106096},
  year    = {2025},
  doi     = {10.1016/j.autcon.2025.106096}
}

@article{yin2026viga,
  title         = {Vision-as-Inverse-Graphics Agent via Interleaved Multimodal Reasoning},
  author        = {Yin, Shaofeng and Ge, Jiaxin and Wang, Zora Zhiruo and Li, Xiuyu and Black, Michael J. and Darrell, Trevor and Kanazawa, Angjoo and Feng, Haiwen},
  journal       = {arXiv preprint arXiv:2601.11109},
  year          = {2026},
  eprint        = {2601.11109},
  archivePrefix = {arXiv},
  primaryClass  = {cs.CV}
}

@inproceedings{xie2025scenescriptrefine,
  title     = {Human-in-the-Loop Local Corrections of 3D Scene Layouts via Infilling},
  author    = {Xie, Christopher and Avetisyan, Armen and Howard-Jenkins, Henry and Siddiqui, Yawar and Straub, Julian and Newcombe, Richard and Balntas, Vasileios and Engel, Jakob},
  booktitle = CVPR,
  year      = {2025}
}

@inproceedings{bieri2025houselayout3d,
  title     = {HouseLayout3D: A Benchmark and Training-Free Baseline for 3D Layout Estimation in the Wild},
  author    = {Bieri, Valentin and Rakotosaona, Marie-Julie and Tateno, Keisuke and Engelmann, Francis and Guibas, Leonidas},
  booktitle = NIPS,
  year      = {2025}
}

@inproceedings{liu2024polyroom,
  title     = {PolyRoom: Room-aware Transformer for Floorplan Reconstruction},
  author    = {Liu, Yuzhou and Zhu, Lingjie and Ma, Xiaodong and Ye, Hanqiao and Gao, Xiang and Zheng, Xianwei and Shen, Shuhan},
  booktitle = ECCV,
  pages     = {343--360},
  year      = {2024},
  publisher = {Springer}
}

@article{sun2025polygraph,
  title   = {PolyGraph: A Graph-Based Method for Floorplan Reconstruction From 3D Scans},
  author  = {Sun, Qian and Fang, Chenrong and Liu, Shuang and Sun, Yidan and Shang, Yu and He, Ying},
  journal = TVCG,
  volume  = {31},
  number  = {10},
  pages   = {7350--7362},
  year    = {2025}
}

@article{ye2025floorsam,
  title         = {FloorSAM: SAM-Guided Floorplan Reconstruction with Semantic-Geometric Fusion},
  author        = {Ye, Han and Wang, Haofu and Zhang, Yunchi and Xiao, Jiangjian and Jin, Yuqiang and Liu, Jinyuan and Zhang, Wen-An and Sychou, Uladzislau and Tuzikov, Alexander and Sobolevskii, Vladislav and Zakharov, Valerii and Sokolov, Boris and Fu, Minglei},
  journal       = {arXiv preprint arXiv:2509.15750},
  year          = {2025},
  eprint        = {2509.15750},
  archivePrefix = {arXiv},
  primaryClass  = {cs.CV}
}

@inproceedings{stekovic2021montefloor,
  title     = {MonteFloor: Extending MCTS for Reconstructing Accurate Large-Scale Floor Plans},
  author    = {Stekovic, Sinisa and Rad, Mahdi and Fraundorfer, Friedrich and Lepetit, Vincent},
  booktitle = CVPR,
  pages     = {16034--16043},
  year      = {2021}
}

@inproceedings{murali2017indoor,
  title     = {Indoor Scan2BIM: Building Information Models of House Interiors},
  author    = {Murali, Srivathsan and Speciale, Pablo and Oswald, Martin R. and Pollefeys, Marc},
  booktitle = {2017 IEEE/RSJ International Conference on Intelligent Robots and Systems},
  pages     = {6126--6133},
  year      = {2017},
  organization = {IEEE},
  doi       = {10.1109/IROS.2017.8206513}
}

@article{ochmann2019volumetric,
  title   = {Automatic Reconstruction of Fully Volumetric 3D Building Models from Oriented Point Clouds},
  author  = {Ochmann, Sebastian and Vock, Richard and Klein, Reinhard},
  journal = {ISPRS Journal of Photogrammetry and Remote Sensing},
  volume  = {151},
  pages   = {251--262},
  year    = {2019},
  doi     = {10.1016/j.isprsjprs.2019.03.017}
}

@inproceedings{wu2023flkpp,
  title     = {Towards Fully Automatic Scan-to-BIM: A Prototype Method Integrating Deep Neural Networks and Architectonic Grammar},
  author    = {Wu, Yijie and Li, Maosu and Xue, Fan},
  booktitle = {Proceedings of the 2023 European Conference on Computing in Construction and the 40th International CIB W78 Conference},
  year      = {2023}
}

@article{han2023floorusg,
  title   = {FloorUSG: Indoor Floorplan Reconstruction by Unifying 2D Semantics and 3D Geometry},
  author  = {Han, Jiali and Liu, Yuzhou and Rong, Mengqi and Zheng, Xianwei and Shen, Shuhan},
  journal = {ISPRS Journal of Photogrammetry and Remote Sensing},
  volume  = {196},
  pages   = {490--501},
  year    = {2023},
}

@article{zbirovsky2025cloud2bim,
  title   = {Open-source Automatic Pipeline for Efficient Conversion of Large-scale Point Clouds to IFC Format},
  author  = {Zbirovsk{\'y}, Sl{\'a}vek and Ne{\v{z}}erka, V{\'a}clav},
  journal = {Automation in Construction},
  volume  = {177},
  pages   = {106303},
  year    = {2025}
}

@article{wong2025imagebased,
  title   = {Image-based Scan-to-BIM for Interior Building Component Reconstruction},
  author  = {Wong, Mun On and Sun, Yifeng and Ying, Huaquan and Yin, Mengtian and Zhou, Hui and Brilakis, Ioannis and Kelly, Tom and Lam, Chi Chiu},
  journal = {Automation in Construction},
  volume  = {173},
  pages   = {106091},
  year    = {2025},
  doi     = {10.1016/j.autcon.2025.106091}
}

@article{jones2020shapeassembly,
  title   = {ShapeAssembly: Learning to Generate Programs for 3D Shape Structure Synthesis},
  author  = {Jones, R. Kenny and Barton, Theresa and Xu, Xianghao and Wang, Kai and Jiang, Ellen and Guerrero, Paul and Mitra, Niloy J. and Ritchie, Daniel},
  journal = TOG,
  volume  = {39},
  number  = {6},
  articleno = {234},
  pages   = {1--20},
  year    = {2020},
  doi     = {10.1145/3414685.3417812}
}

@inproceedings{ye2025primitiveanything,
  title     = {PrimitiveAnything: Human-Crafted 3D Primitive Assembly Generation with Auto-Regressive Transformer},
  author    = {Ye, Jingwen and He, Yuze and Zhou, Yanning and Zhu, Yiqin and Xiao, Kaiwen and Liu, Yong-Jin and Yang, Wei and Han, Xiao},
  booktitle = {ACM SIGGRAPH 2025 Conference Papers},
  year      = {2025},
  doi       = {10.1145/3721238.3730732}
}

@article{raji2026proc3d,
  title         = {Proc3D: Procedural 3D Generation and Parametric Editing of 3D Shapes with Large Language Models},
  author        = {Raji, Fadlullah and Petrangeli, Stefano and Gadelha, Matheus and Shen, Yu and Bhattacharya, Uttaran and Wu, Gang},
  journal       = {arXiv preprint arXiv:2601.12234},
  year          = {2026},
  eprint        = {2601.12234},
  archivePrefix = {arXiv},
  primaryClass  = {cs.GR}
}

@inproceedings{feng2025casagpt,
  title     = {CASAGPT: Cuboid Arrangement and Scene Assembly for Interior Design},
  author    = {Feng, Weitao and Zhou, Hang and Liao, Jing and Cheng, Li and Zhou, Wenbo},
  booktitle = CVPR,
  year      = {2025}
}

@inproceedings{hu2024scenecraft,
  title     = {SceneCraft: An LLM Agent for Synthesizing 3D Scene as Blender Code},
  author    = {Hu, Ziniu and Iscen, Ahmet and Jain, Aashi and Kipf, Thomas and Yue, Yisong and Ross, David A. and Schmid, Cordelia and Fathi, Alireza},
  booktitle = {Proceedings of the International Conference on Machine Learning},
  year      = {2024}
}

@article{yao2026photoagent,
  title         = {PhotoAgent: Agentic Photo Editing with Exploratory Visual Aesthetic Planning},
  author        = {Yao, Mingde and You, Zhiyuan and Tam, King-Man and Wang, Menglu and Xue, Tianfan},
  journal       = {arXiv preprint arXiv:2602.22809},
  year          = {2026},
  eprint        = {2602.22809},
  archivePrefix = {arXiv},
  primaryClass  = {cs.CV}
}

@article{du2026text2bim,
  title   = {Text2BIM: Generating Building Models Using a Large Language Model-Based Multiagent Framework},
  author  = {Du, Changyu and Esser, Sebastian and Nousias, Stavros and Borrmann, Andr{\'e}},
  journal = {Journal of Computing in Civil Engineering},
  volume  = {40},
  number  = {2},
  pages   = {04025142},
  year    = {2026},
  doi     = {10.1061/JCCEE5.CPENG-6386}
}

@article{nithyanantham2025mcp4ifc,
  title         = {MCP4IFC: IFC-Based Building Design Using Large Language Models},
  author        = {Nithyanantham, Bharathi Kannan and Sesterhenn, Tobias and Nedungadi, Ashwin and Garijo, Sergio Peral and Zenkner, Janis and Bartelt, Christian and L{\"u}dtke, Stefan},
  journal       = {arXiv preprint arXiv:2511.05533},
  year          = {2025},
  eprint        = {2511.05533},
  archivePrefix = {arXiv},
  primaryClass  = {cs.CL}
}

@article{deng2025bimgent,
  title   = {BIMgent: Towards Autonomous Building Modeling via Computer-use Agents},
  author  = {Deng, Zihan and Du, Changyu and Nousias, Stavros and Borrmann, Andr{\'e}},
  journal = {ICML 2025 Workshop on Computer Use Agents},
  year    = {2025}
}

@article{liu2025cage,
  title         = {CAGE: Continuity-Aware edGE Network Unlocks Robust Floorplan Reconstruction},
  author        = {Liu, Yiyi and Liu, Chunyang and Wang, Bohan and Jiao, Weiqin and Wu, Bojian and Fan, Lubin and Chen, Yuwei and Li, Fashuai and Xiong, Biao},
  journal       = {arXiv preprint arXiv:2509.15459},
  year          = {2025},
  eprint        = {2509.15459},
  archivePrefix = {arXiv},
  primaryClass  = {cs.CV},
  doi           = {10.48550/arXiv.2509.15459}
}

@inproceedings{jia2020structured3d,
  title     = {Structured3D: A Large Photo-realistic Dataset for Structured 3D Modeling},
  author    = {Jia Zheng and Junfei Zhang and Jing Li and Rui Tang and Shenghua Gao and Zihan Zhou},
  booktitle = ECCV,
  year      = {2020}
}

@article{khoshelham2017isprs,
  title={The ISPRS benchmark on indoor modelling},
  author={Khoshelham, Kourosh and D{\'\i}az Vilari{\~n}o, Lucia and Peter, Michael and Kang, Zhizhong and Acharya, Debaditya},
  journal=ISPRS,
  volume={42},
  pages={367--372},
  year={2017},
  publisher={Copernicus GmbH}
}

@InProceedings{avetisyan2020scenecad,
    title={SceneCAD: Predicting Object Alignments and Layouts in RGB-D Scans},
    author={Armen Avetisyan and Tatiana Khanova and Christopher Choy and Denver Dash and Angela Dai and Matthias Nießner},
    booktitle = ECCV,
    month={August},
    year={2020},
}

@inproceedings{matt2022procthor,
  author={Matt Deitke and Eli VanderBilt and Alvaro Herrasti and
          Luca Weihs and Jordi Salvador and Kiana Ehsani and
          Winson Han and Eric Kolve and Ali Farhadi and
          Aniruddha Kembhavi and Roozbeh Mottaghi},
  title={{ProcTHOR: Large-Scale Embodied AI Using Procedural Generation}},
  booktitle=NIPS,
  year={2022},
  note={Outstanding Paper Award}
}

@inproceedings{khanna2024hssd,
  title={Habitat synthetic scenes dataset (hssd-200): An analysis of 3d scene scale and realism tradeoffs for objectgoal navigation},
  author={Khanna, Mukul and Mao, Yongsen and Jiang, Hanxiao and Haresh, Sanjay and Shacklett, Brennan and Batra, Dhruv and Clegg, Alexander and Undersander, Eric and Chang, Angel X and Savva, Manolis},
  booktitle=CVPR,
  pages={16384--16393},
  year={2024}
}

@inproceedings{wu2025sonata,
    title={Sonata: Self-Supervised Learning of Reliable Point Representations},
    author={Wu, Xiaoyang and DeTone, Daniel and Frost, Duncan and Shen, Tianwei and Xie, Chris and Yang, Nan and Engel, Jakob and Newcombe, Richard and Zhao, Hengshuang and Straub, Julian},
    booktitle=CVPR,
    year={2025}
}

@article{an2024qwen,
    title   = {Qwen2.5 Technical Report}, 
    author  = {An Yang and Baosong Yang and Beichen Zhang and Binyuan Hui and Bo Zheng and Bowen Yu and Chengyuan Li and Dayiheng Liu and Fei Huang and Haoran Wei and Huan Lin and Jian Yang and Jianhong Tu and Jianwei Zhang and Jianxin Yang and Jiaxi Yang and Jingren Zhou and Junyang Lin and Kai Dang and Keming Lu and Keqin Bao and Kexin Yang and Le Yu and Mei Li and Mingfeng Xue and Pei Zhang and Qin Zhu and Rui Men and Runji Lin and Tianhao Li and Tingyu Xia and Xingzhang Ren and Xuancheng Ren and Yang Fan and Yang Su and Yichang Zhang and Yu Wan and Yuqiong Liu and Zeyu Cui and Zhenru Zhang and Zihan Qiu},
    journal = {arXiv},
    year    = {2024}
}

@inproceedings{kulikov2025flowedit,
  title={Flowedit: Inversion-free text-based editing using pre-trained flow models},
  author={Kulikov, Vladimir and Kleiner, Matan and Huberman-Spiegelglas, Inbar and Michaeli, Tomer},
  booktitle=ICCV,
  pages={19721--19730},
  year={2025}
}

@inproceedings{gong2025instantedit,
  title        = {InstantEdit: Text-Guided Few-Step Image Editing with Piecewise Rectified Flow},
  author       = {Gong, Yiming and Zhu, Zhen and Zhang, Minjia},
  booktitle    = ICCV,
  year         = {2025}
}
